\documentclass[10pt,twocolumn,letterpaper]{article}

\usepackage{wacv}
\usepackage{times}
\usepackage{epsfig}
\usepackage{graphicx}
\usepackage{amsmath, makecell}
\usepackage{amssymb}
\usepackage{booktabs}

\usepackage[accsupp]{axessibility}
\ifwacvfinal
\usepackage[breaklinks=true,colorlinks,bookmarks=false]{hyperref}
\else
\usepackage[pagebackref=true,breaklinks=true,colorlinks,bookmarks=false]{hyperref}
\fi

\usepackage[capitalize]{cleveref}

\usepackage{booktabs,multicol,multirow,comment,xspace}
\usepackage[inline]{enumitem}

\usepackage{subfigure}

\usepackage{xcolor,soul}
\definecolor{todocolor}{RGB}{200,120,120}
\sethlcolor{todocolor}

\usepackage{pifont}
\newcommand{\cmark}{\ding{51}}%
\newcommand{\xmark}{\ding{55}}%

\newcommand{\myfirstpara}[1]{\par \noindent \textbf{#1:}}
\newcommand{\mypara}[1]{\vspace{0.5em} \myfirstpara{#1}}

\newcommand\iou{\texttt{IoU}\xspace}

\newcommand{\tss}{\texttt{S3Net}\xspace}
\newcommand\sota{\texttt{SOTA}\xspace}
\newcommand\nms{\texttt{NMS}\xspace}
\newcommand\rpn{\texttt{RPN}\xspace}
\newcommand\msma{\texttt{MSMA}\xspace}
\newcommand\evs{\texttt{EVS}\xspace}
\newcommand\nli{\texttt{NLI}\xspace}
\newcommand\evda{\texttt{EV17}\xspace}
\newcommand\evdb{\texttt{EV18}\xspace}

\newcommand\isinet{\texttt{ISINet}\xspace}
\newcommand\trasetr{\texttt{TraSeTR}\xspace}

\graphicspath{{../}}

\def\wacvPaperID{869} 

\wacvapplicationstrack 

\wacvfinalcopy 

\crefname{section}{Sec.}{Secs.}
\Crefname{section}{Section}{Sections}
\Crefname{table}{Table}{Tables}
\crefname{table}{Tab.}{Tabs.}
\begin{document}

\title{%
	From Forks to Forceps: A New Framework for  \\ 
	Instance Segmentation of Surgical Instruments %
}

\author{Britty Baby$^{1,2}$, Daksh Thapar$^{3}$, Mustafa Chasmai$^{1}$, Tamajit Banerjee$^{1}$, Kunal Dargan$^{1}$, \\
Ashish Suri$^{2,1}$, Subhashis Banerjee$^{4,1}$, Chetan Arora$^{1}$\\
$^{1}$IIT Delhi, India, 
$^{2}$AIIMS, New Delhi, India, $^{3}$IIT Mandi, India, $^{4}$Ashoka University, India
}

\maketitle
\thispagestyle{empty}

\begin{abstract}
Minimally invasive surgeries and related applications demand surgical tool classification and segmentation at the instance level. Surgical tools are similar in appearance and are long, thin, and handled at an angle. The fine-tuning of state-of-the-art (\sota) instance segmentation models trained on natural images for instrument segmentation has difficulty discriminating instrument classes. Our research demonstrates that while the bounding box and segmentation mask are often accurate, the classification head misclassifies the class label of the surgical instrument. We present a new neural network framework that adds a classification module as a new stage to existing instance segmentation models. This module specializes in improving the classification of instrument masks generated by the existing model. The module comprises multi-scale mask attention, which attends to the instrument region and masks the distracting background features. We propose training our classifier module using metric learning with \emph{arc loss} to handle low inter-class variance of surgical instruments. We conduct exhaustive experiments on the benchmark datasets EndoVis2017 and EndoVis2018. We demonstrate that our method outperforms all (more than 18) \sota methods compared with, and improves the \sota performance by at least 12 points (20\%) on the EndoVis2017 benchmark challenge and generalizes effectively across the datasets. Project page with source code is available at \url{nets-iitd.github.io/s3net}.

\end{abstract}

\section{Introduction}
\label{sec:introduction}


\begin{figure}[t]
\centering
\def\imgwidth{0.3\linewidth}
\subfigure[GT]{
	\includegraphics[width=\imgwidth]{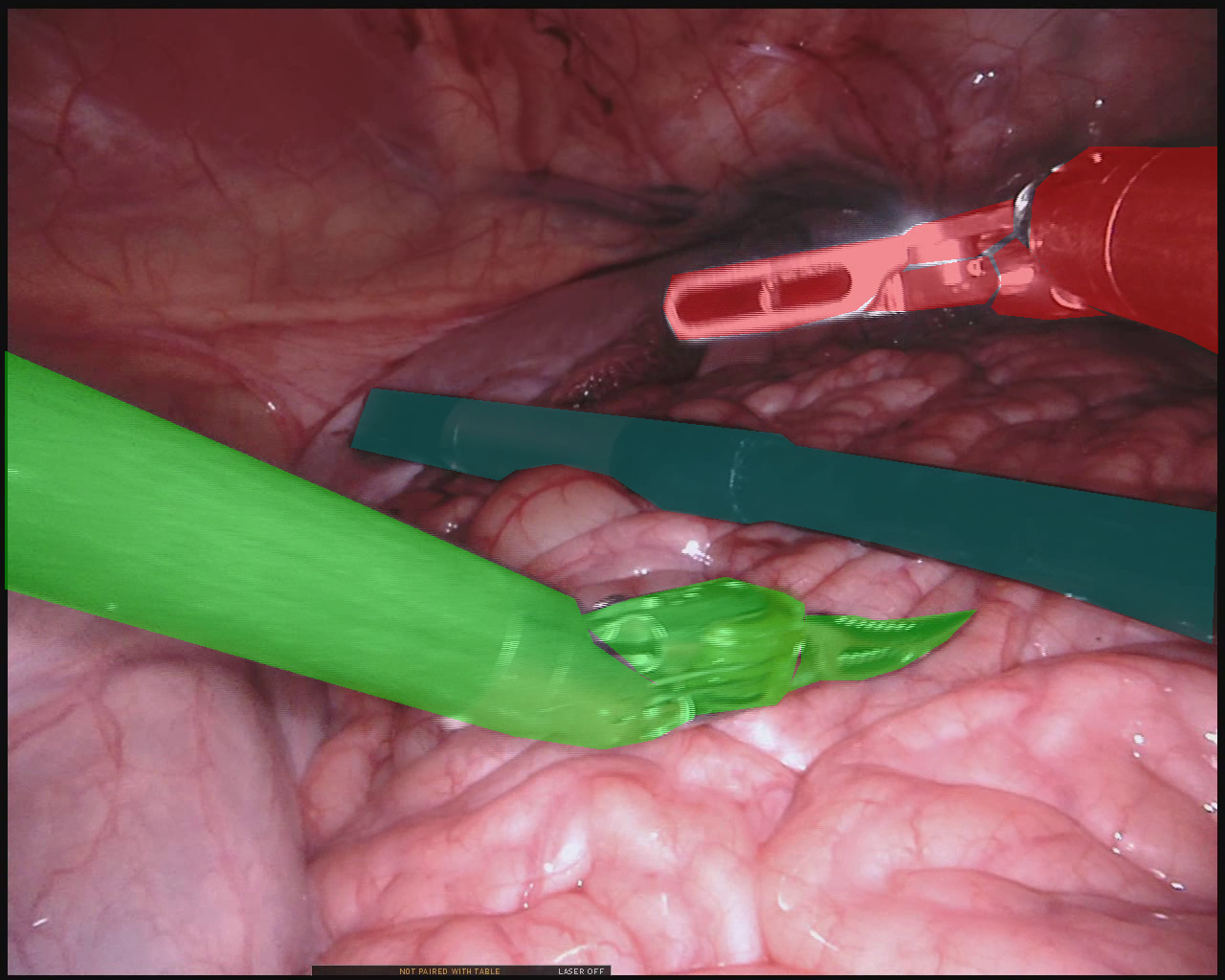}
}
\subfigure[\cite{TernausNet}]{
	\includegraphics[width=\imgwidth]{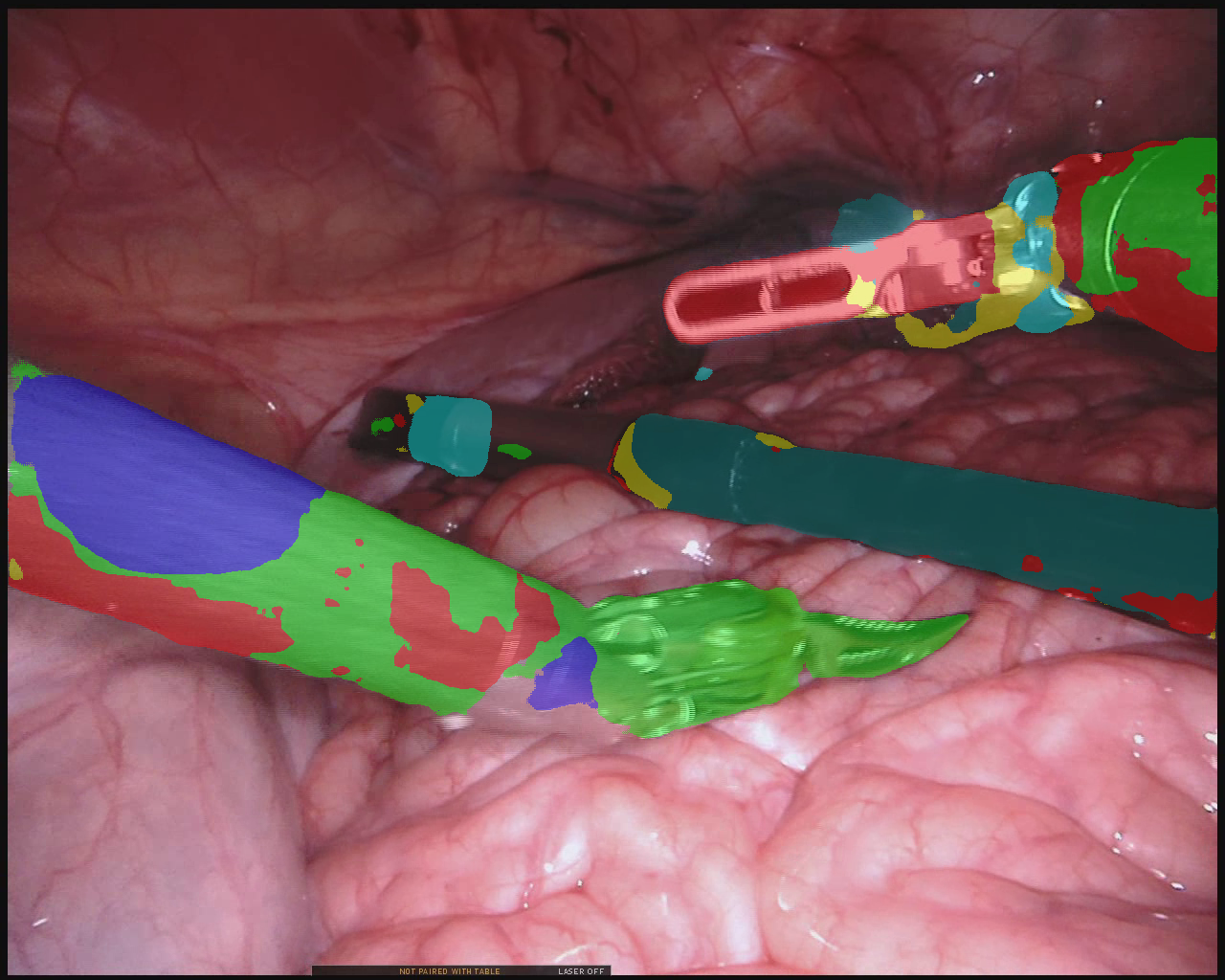}
}
\subfigure[\cite{MaskRCNN}]{
	\includegraphics[width=\imgwidth]{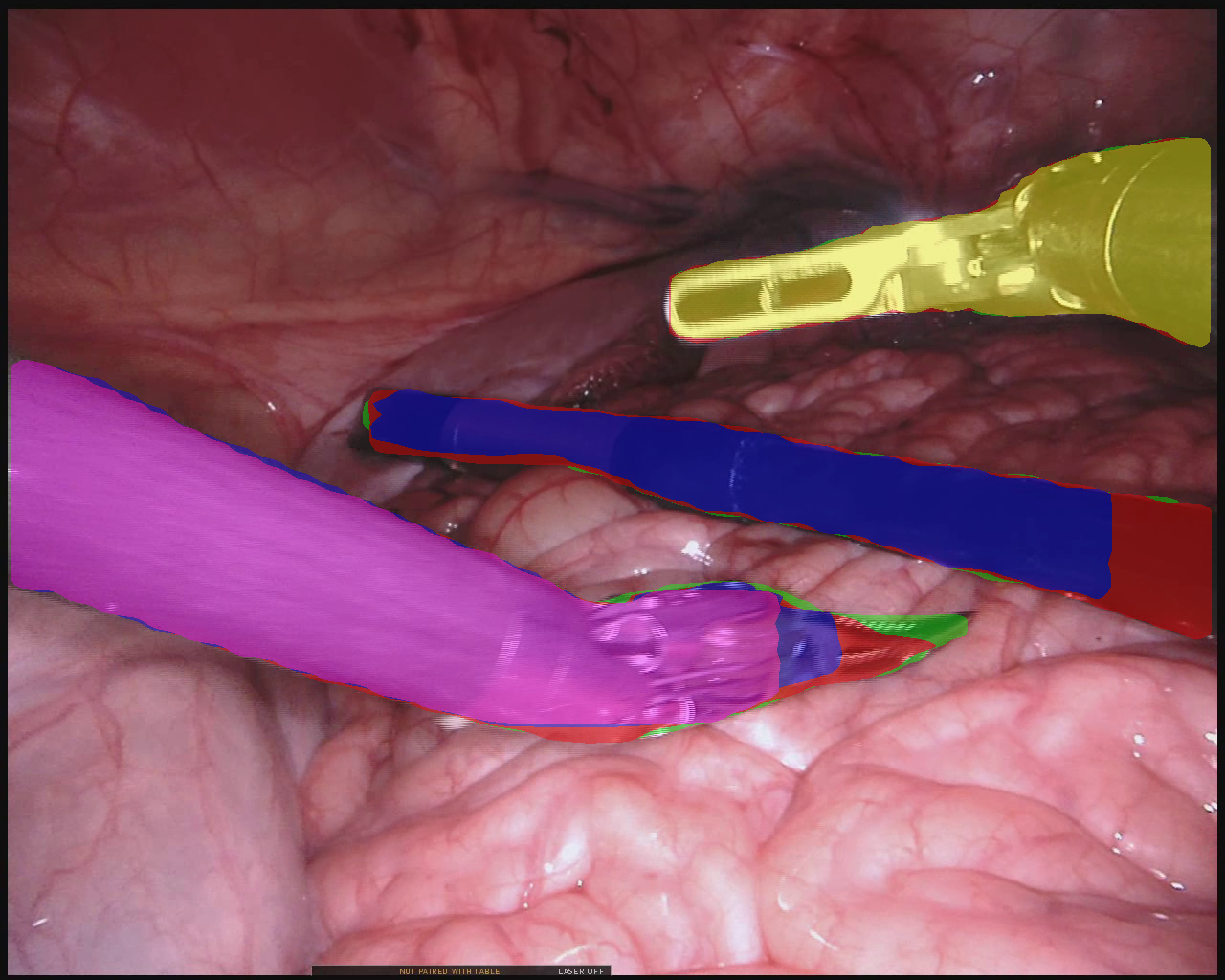}
}
\subfigure[\cite{gonzalez2020isinet}]{
	\includegraphics[width=\imgwidth]{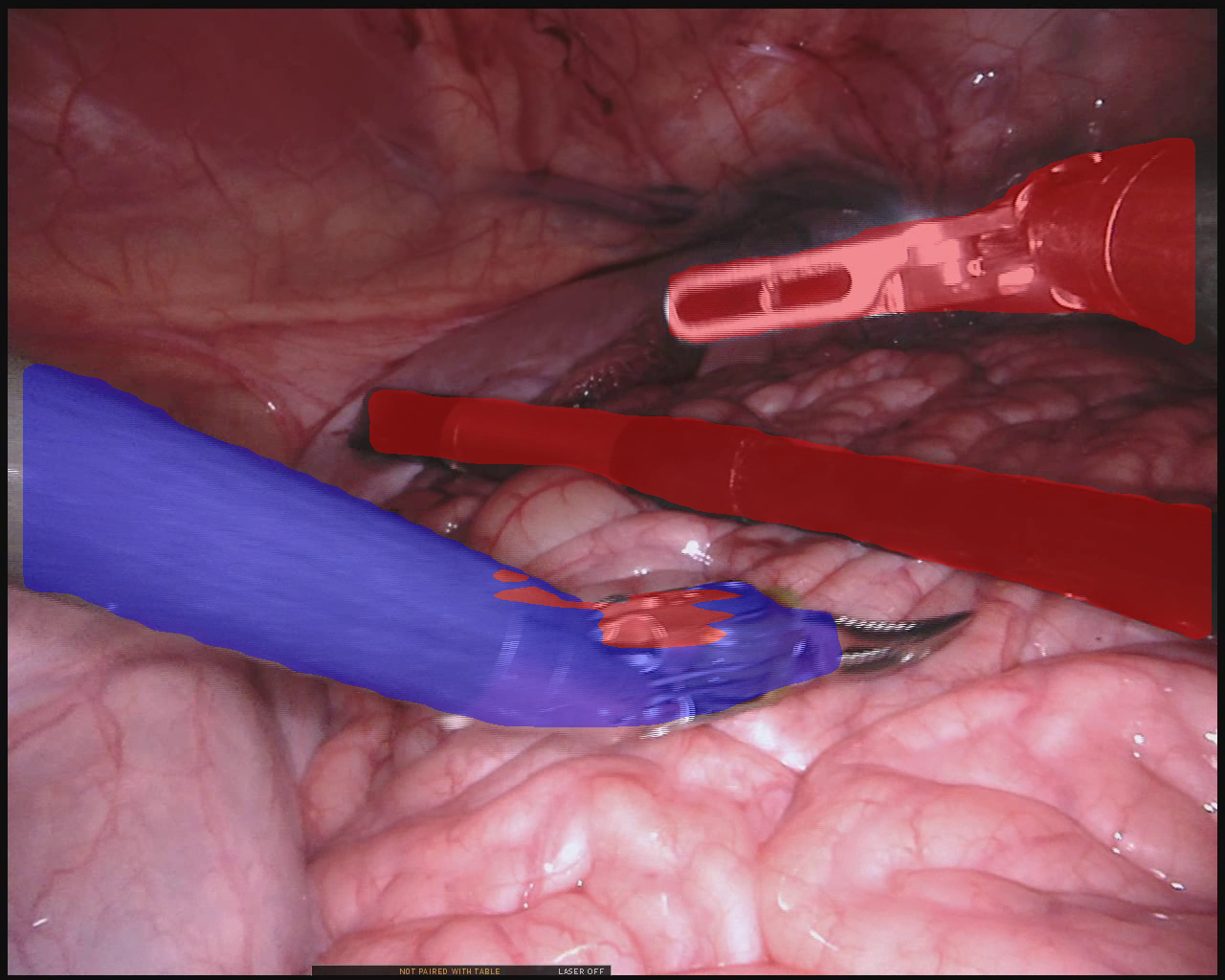}
} 
\subfigure[\cite{DetectoRS}]{
	\includegraphics[width=\imgwidth]{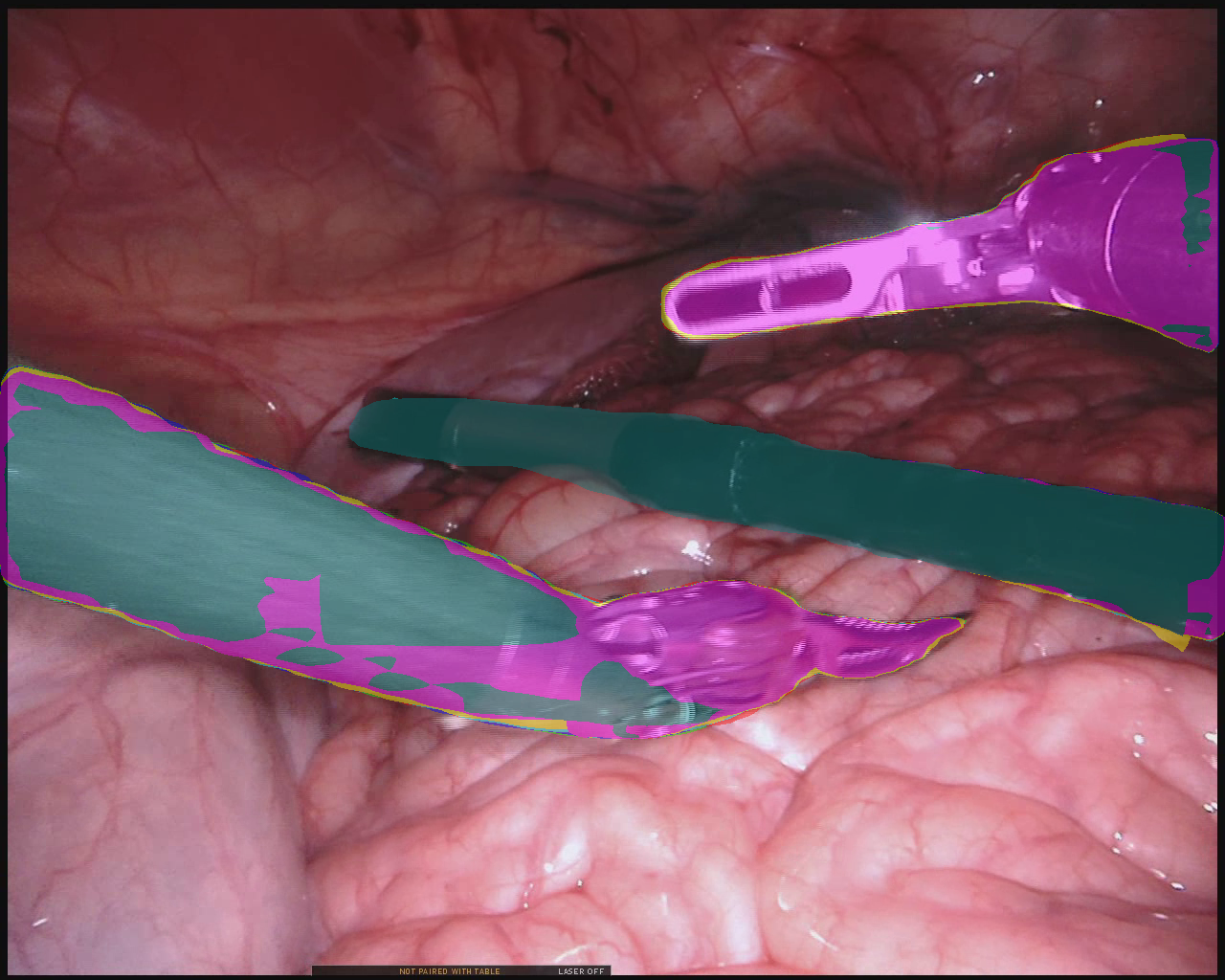}
} 
\subfigure[Ours]{
	\includegraphics[width=\imgwidth]{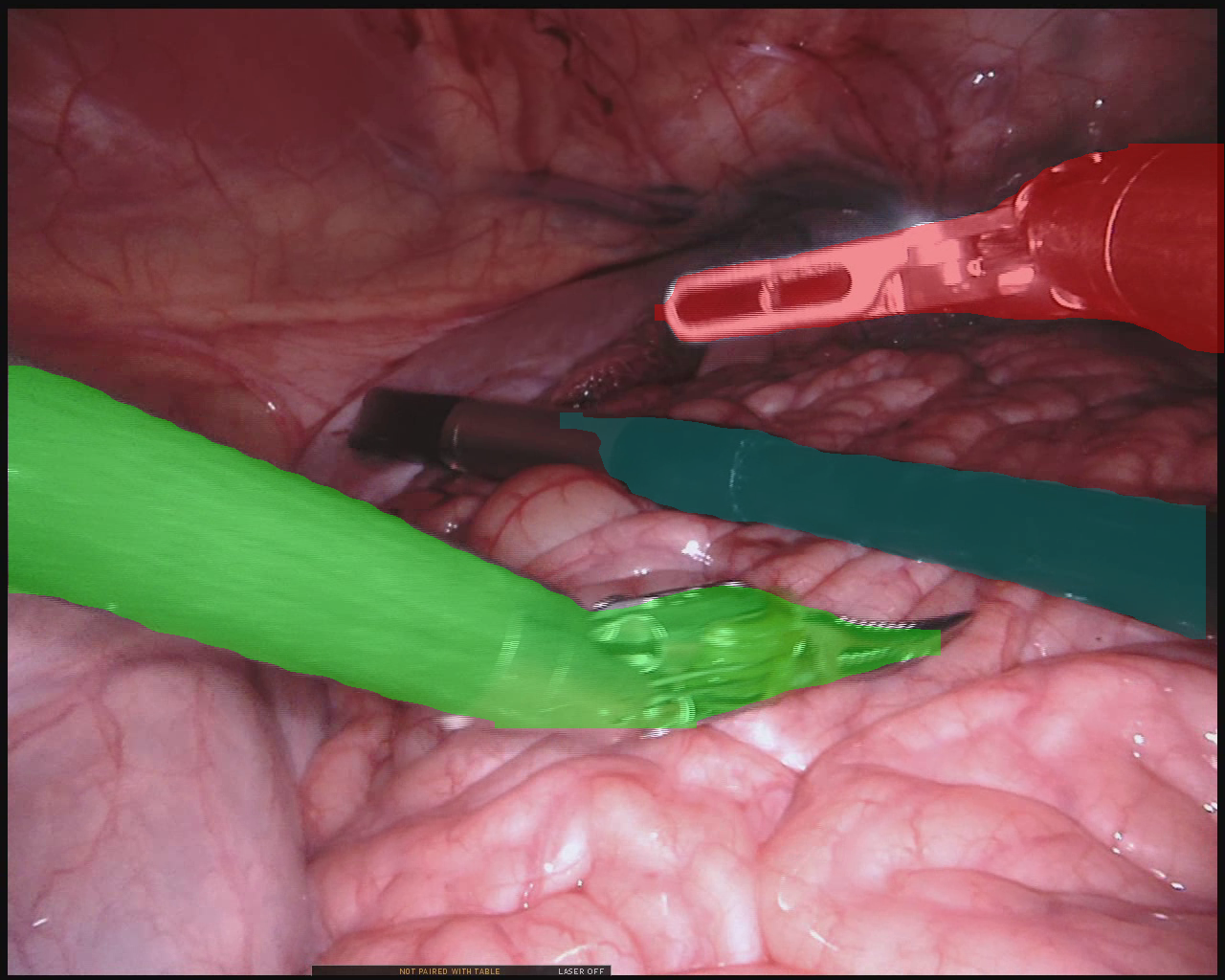}
}
\caption{Instrument segmentation produced by various competitive methods on a sample from the \evda dataset~\cite{EndoVis_2017}. Each instrument class is shown in a different color. Note that ISINet~\cite{gonzalez2020isinet} gets the segmentation right but classifies incorrectly. We identify instrument misclassification as the primary reason for low performance of the \sota techniques, and propose various architecture modification for accurate classification. To illustrate the severity of the problem, substituting the predicted class label of a MaskRCNN object with the ground truth label improved the model's AP50 score from 0.65 to 0.90.}
\label{fig:Teaser}
\end{figure}
%


The computer vision community has significantly progressed in designing semantic and instance segmentation algorithms in recent years. One of the reasons for the success is the availability of large  datasets~\cite{russakovsky2015imagenet, COCO_dataset, Imagenet_dataset, cityscapes}. On the other hand, due to the advantages of small incisions and rapid recovery, minimally invasive surgeries (MIS) are increasingly accepted in various surgical specialties~\cite{MIS_benefits_2, MIS_benefits_1}. Automated segmentation of a surgical instrument in MIS is an area of active research with high utility. The surgical instrument segmentation poses various challenges depending on the dataset acquisition source, type of surgery and instruments/ tools involved, image resolution, dataset size, tool statistics, challenging conditions (occlusions, rapid appearance changes, specular reflections, smoke, blur, blood spatter)~\cite{visionreference}. 

Most surgical datasets and algorithms structure instrument segmentation as semantic segmentation, which classifies each pixel as one of the instrument class~\cite{TernausNet, hasan2019u, islam2019real, MFTAPNet, PAANet, DMNet}. Due to disconnected regions and occluded/overlapping instruments, the task of assigning an instance label to the semantic segmentation output is non-trivial. However, obtaining an instance-level mask of the manipulating instruments is essential for most surgical instrument segmentation applications that depend on instrument tracking \cite{du2018articulated, kurmann2017simultaneous, sznitman2014fast, jin2018tool, baby2016neuro}. Hence, we argue for formulating the task as multi-class instance segmentation. 

The research in surgical instrument segmentation is largely driven by the EndoVis2017 dataset~\cite{EndoVis_2017}, which is a robotic instrument dataset containing annotations for different instrument types. The dataset contains seven instruments, all of which have a thin, long, tube-like structure. ISINet \cite{du2018articulated} fine-tunes a MaskRCNN \cite{MaskRCNN} backbone for instance segmentation with a reported Challenge \iou score of 0.55. TraSeTR \cite{TraSeTR} uses a transformer architecture that exploits tracking cues to assist surgical instrument segmentation with a Challenge \iou score of 0.6.

\mypara{Contribution 1}
We investigated the reasons for low \iou scores of \sota algorithms on medical instrument segmentation. We found that these methods give a reasonable output for the bounding box and segmentation mask but often misclassify the output box/mask (\cref{fig:Teaser}). We believe that our observation is analogous to the one reported by~\cite{wang2020devil} for natural images. The authors have reported that in the dataset with a long tail, the \sota techniques for object detection in natural images give correct region proposals for less frequent classes but often misclassify them. We posit that due to significant visual differences between natural objects and medical instruments, a deep neural network model that does cross-domain fine-tuning is unable to develop robust features for classification. However, since the bounding box and mask predictions are based on more robust features such as edges, these predictions generalize more easily. Therefore, there is a need for a specialized module in these techniques that focuses on acquiring the attributes necessary for efficiently classifying surgical instruments. Hence, we propose adding a dedicated classification module as a new stage in the existing techniques, which decouples the classification from the bounding box and mask prediction and specializes in classifying classes from the tail of a distribution.

\mypara{Contribution 2}
A deeper investigation found a variation in aspect ratio and orientation between natural images and MIS. While in natural images, the width-to-height ratio is usually around 0.5, surgical instruments are mostly two or greater. Further, natural objects appear mostly vertical in an image and fit well in rectilinear bounding boxes. On the other hand, surgical instruments are used obliquely and appear across a bounding box's diagonal. The instrument's aspect ratio and oblique appearance reduce its proportion in the bounding box area and brings in a distracting background.
To make matters worse, given the small operating regions in an MIS, a proposed bounding box may contain multiple tools, further complicating the classification task. The finding motivates the need for classification based on mask-based attention rather than the existing bounding box-based one. Hence, we propose to include mask-based attention in the proposed specialized classification module.

\mypara{Contribution 3}
Surgical instruments show inter-class appearance similarity and contain long shafts; the only distinguishing characteristic may be the instruments' tips. Therefore, generic cross entropy-based training of classifiers in contemporary architecture is unsuitable for fine-grained classification of surgical tools. Recent literature suggests that for small datasets, it is beneficial to separate representation learning and classification stages \cite{yao2020adaptive}. The first can be achieved using a contrastive loss, followed by fine-tuning for the classification. We follow a similar approach and train our proposed classification module using arc loss \cite{deng2019arcface}, followed by fine-tuning with cross-entropy loss.

\mypara{Results}
We conduct exhaustive experiments on the benchmark Robot-assisted surgery datasets EndoVis2017 (\evda), and EndoVis2018 (\evdb).
The proposed method generalizes well on all these datasets, outperforms the instance segmentation methods with varying backbones, and achieves at least 12 points (20\%) improvement over the \sota on the benchmark EndoVis2017 challenge.
%

\begin{figure*}[t]
	\centering
	\includegraphics[width=\linewidth]{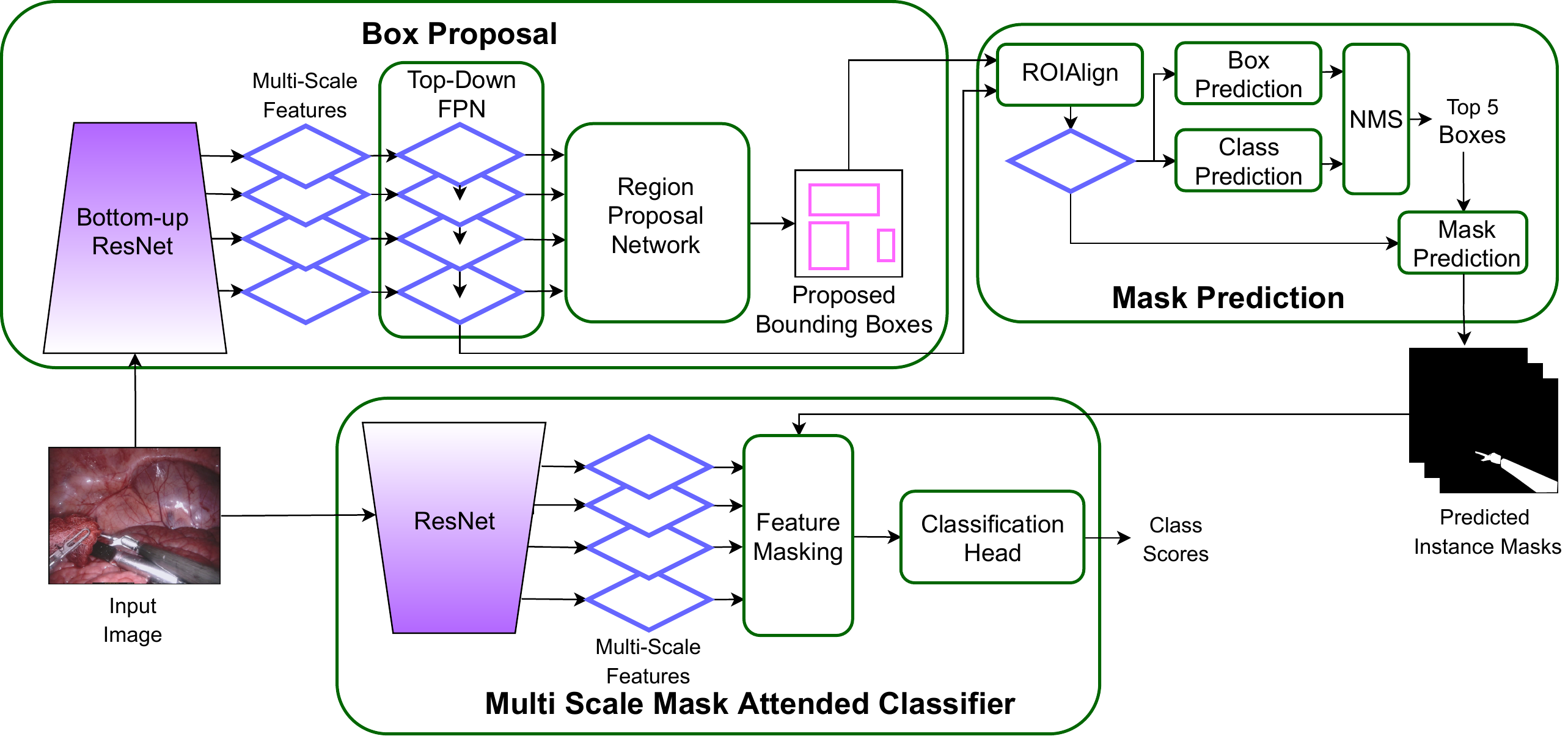}
	\caption{Architecture of the proposed 3-stage neural network model, named \tss, for the instrument segmentation. Whereas the first two stages are similar to the state of the art, we introduce a third stage, named \msma, specializing in classification. We make several innovations in the design of \msma as described in the main text.}
	\label{fig:arch}
\end{figure*}
\section{Related Work}
\label{sec:Related Work}

The application of object detection, segmentation, and tracking in the field of MIS extends to various surgical branches like gynecology, ophthalmology, and neurosurgery~\cite{bodenstedt2018comparative, zadeh2020surgai, ni2019raunet, qin2020towards}. Researchers have contributed different datasets in this regard as well~\cite{EndoVis, ross2020robust, cadis}. Many techniques have been developed using the Endovis challenge datasets. The instrument segmentation problem has been formulated using both semantic~\cite{TernausNet, hasan2019u, islam2019real, MFTAPNet, PAANet, DMNet}, as well as instance segmentation ~\cite{MaskRCNN_by_Anchor_Opt, gonzalez2020isinet, islam2020ap, Mask_then_classify}. The approach for this segmentation can be supervised, semi-supervised or unsupervised methods. The semi-supervised/unsupervised methods handle the data annotation scarcity in the medical domain and explore domain adaptation of the model to surgical scenario ~\cite{sim2real_sahu, liu2020unsupervised_binary, unsupervised_MICCAI20, unsupervised_MICCAI21}. In this work, we are focusing on the supervised instance segmentation problem. 

\mypara{Semantic Segmentation of Medical Instruments}
TernausNet uses U-Net architecture ~\cite{UNet}, on a pre-trained VGG11 or VGG16 backbone~\cite{TernausNet}. It shows the best performance on binary segmentation but performs poorly on the classification of the instrument type. U-NetPlus uses a modified Encoder-Decoder-based U-Net architecture and data augmentation techniques to improve performance~\cite{hasan2019u}. Some methods explore real-time instrument semantic segmentation~\cite{islam2019real, DMNet}. PAANet aggregate multi-scale attentive features ~\cite{PAANet} and MF-TAPNet integrates flow-based temporal priors to an attention pyramid network~\cite{MFTAPNet}. All the methods discussed above use a single-stage approach for semantic segmentation, which often over segments an instrument to multiple classes (see \cite{TernausNet} in \cref{fig:Teaser}).

\mypara{Instance Segmentation of Medical Instruments}
The formulation has been explored in two ways: cross-domain fine-tuning of models pre-trained on natural images~\cite{MaskRCNN, MaskRCNN_by_Anchor_Opt, gonzalez2020isinet, TraSeTR}, and custom-designed models for the task~\cite{islam2020ap, Mask_then_classify}. For the first category, researchers have primarily used MaskRCNN~\cite{MaskRCNN}. Kong et al.~\cite{MaskRCNN_by_Anchor_Opt} adapted MaskRCNN by optimizing the anchor scales for instrument types. ISINet~\cite{gonzalez2020isinet} has used fine-tuned MaskRCNN along with a temporal consistency module to exploit the sequential nature of the data. The improved performance in ISINet is due to the non-maximum suppression of regions across various classes and retaining only the highest predicted class for any instrument instance. Their temporal consistency module improves marginally over their instance selection heuristic. TraSeTR~\cite{TraSeTR} is a transformer-based track-to-segment method that incorporates tracking cues for instance segmentation of instruments. This technique relies on the second stage's classification predictions and adds identity matching and contrastive query learning to address surgical instruments with huge temporal variations. AP-MTL~\cite{islam2020ap} has proposed an Encoder-Decoder architecture for real-time instance segmentation. They have shown improvement over a domain-adapted MaskRCNN. Mask-then-classify~\cite{Mask_then_classify} also used an encoder-decoder network, along with a classifier that uses features from the segmentation stage to classify the pixel-wise instances. The approach uses a single-stage network and is prone to classification errors if there are errors in the masks and vice-versa. 

In this work, we focus on the misclassification challenges of the bounding-box-based instance segmentation methods when fine-tuned for instrument segmentation. We propose adding a novel specialized classification module to mitigate the challenges.

\section{Proposed Architecture}
\label{sec:Our_Proposal}

As such, our main contribution, the specialized classification module as a new stage, can be inserted into any existing instance segmentation model. However, we have based our model on the MaskRCNN \cite{MaskRCNN} backbone for validation. The MaskRCNN backbone contains two stages corresponding to the region proposal network (RPN) and a classification head generating masks and labels for each proposal. We insert our classification module as the third stage in the MaskRCNN and replace the labels generated by the second stage with the ones generated by our module. The architecture of our proposed Three Stage Deep Neural Network (\tss) is shown in \cref{fig:arch}. 

\mypara{Notation}
For a given input frame $I_i$, the first stage, (Box Proposal) extracts $l$ bounding box proposals, where $B_{i,j}$ is the $j^\text{th}$ proposal in the $i^\text{th}$ frame. The second stage (Mask Prediction), predicts the mask $P_{i,j,\hat{c}}$ for each instrument using the bounding box proposals. Here, $\hat{c}$ refers to the class predicted at this stage. 

\mypara{Processing of second stage output}
The first stage of MaskRCNN and other similar models is called Region Proposal Network (\rpn) and typically outputs many overlapping regions corresponding to a single object instance in the image. The classification head of the MaskRCNN remains weak even after cross-domain fine-tuning. Hence, many of these overlapping boxes get classified as different classes. A typical non-maximal suppression (\nms) step does not reject overlapping boxes corresponding to different classes. If not addressed, this leads to many false-positive predictions by MaskRCNN and other \sota techniques that we have compared within our experiments. Hence, we modify our implementation's standard \nms step to reject overlapping segments in an image across the classes.

\mypara{Handling misclassification}
In a two-stage network, we observe that the proposals generated by the first stage \rpn are inaccurate. However, these proposals get refined by the bounding box regression head in the second stage, leading to higher bounding box and mask accuracy after the second stage. However, the classification is performed on the regions cropped out of inaccurate region proposals from the first stage and remains fragile, which makes classification the bottleneck in the instrument segmentation accuracy. Based upon the insight gained from our analysis, we propose a new deep neural network paradigm that uses the first two stages from a standard instance segmentation method but contains an additional third stage specializing in classification based on the masks. We call the proposed classifier as \emph{Multi-Scale Mask Attended Classifier (\msma)}, which updates/corrects the class predictions from the first two stages. Let $\hat{c}$ denote the class label, and $P_{i,j,\hat{c}}$ denote the mask output corresponding to a region proposal. Then the objective of \msma is to take the original image and mask $P_{i,j,\hat{c}}$ as the input and refine the class label from $\hat{c}$ to a more accurate label $c$. The final mask (with updated class label) is denoted as $P_{i,j,c}$. 
As described earlier, the bounding box rectangular regions for a medical instrument contain lots of background and the pixels of other instruments. This is due to the instrument shape and how instruments are typically used in a surgery. This distracts the classification head and leads to errors. Hence, instead of using rectangular region proposals, we introduce spatial mask attention in \msma to emphasize the region belonging to the instrument only. 


During training, we use ground truth masks corresponding to the instance, and while testing, we use the mask predicted by the second stage. This hard-mask attention is performed on multi-scale features of the image. This helps our model focus on the correct instrument spatial region in the image, leading to more accurate classification of the mask generated in the second stage of MaskRCNN. Further, to effectively train with a small dataset, we separate learning feature representation and classification in the proposed third stage. We first perform metric learning with arc loss function, followed by a learning classifier with categorical cross-entropy. Below we describe the proposed \msma module.

\mypara{Multi Scale Mask Attended (\msma) Classifier}
Convolutional Feature masking ~\cite{convfeaturemasking} was proposed by Dai et al. to exploit shape information to separate objects from the stuff. We adopt this in an instance segmentation framework to separate instruments from the background/ overlapping instruments and improve the classifier. We explore the decoupling property of the mask and the classification head and use a dedicated neural network with multi-scale mask attention for classification. Our proposed paradigm is shown in \cref{fig:arch}. It takes as input the original RGB image $I_i$ and the predicted masks $P_{i,j,\hat{c}}$ of each instrument instance. A ResNet~\cite{he2016deep} backbone is used to extract multi-scale features from $I_i$. 
%
%
Then, the mask $P_{i,j,\hat{c}}$ is multiplied by each feature to create multi-scale mask-attended features. The masked features are then merged using another $1\times 1$ convolution, creating a single feature map for each instance. Note that if multiple instances of a class are predicted in a frame, then the \msma classifier is run separately for each instance.

%

We learn an embedding layer over the masked feature map, which outputs an embedding, $E_{i,j}$, for each instrument instance. Each $E_{i,j}$ is then used to classify the instrument present in the mask, giving us a new class label $c$ for the mask. 
For training \msma classifier, we utilize arc loss~\cite{deng2019arcface}, as defined below:
\begin{equation*}
	\mathcal{L} = -\frac{1}{C} \sum_{c=1}^{c=C} \log \frac{e^{\cos(\theta_{c}+m)}}{e^{\cos(\theta_{c}+m)} + \sum_{j=1,j\neq c}^{C} e^{\cos \theta_j}}.
\end{equation*} 
Here $C$ is the number of classes, and $m$ is the angular margin enforced between features of different classes. Further, $\theta_{j}$ is the angle formed between the Embedding feature $E_j$ and the weight vector of the $j^\text{th}$ neuron in the final fully connected layer. The arc loss is adapted from the face recognition domain to the surgical domain where the inter-class variance is low; the arc loss tries to maximize the distance between the features of the classes, thereby increasing the classification accuracy. Unlike categorical cross-entropy loss, which computes the dot product between $E_{i,j}$ and each weight vector, the arc loss only depends on the angle between them. Using arc loss removes the effect of the magnitude of the weight vector for the final decision. Since the magnitude of weight vectors is unbounded, they can easily become biased for a class with more samples. Hence, the arc loss handles class imbalance in the data by removing the dependency over the magnitude of weight vectors. Moreover, the arc loss forms a metric-based angular cluster for each class rather than learning a decision boundary between various classes. This is ensured by the angular margin $m$, resulting in better intra-class compactness and inter-class separability despite data scarcity.

\section{Dataset and Evaluation}

\myfirstpara{Benchmark Datasets}
We have used Robot-assisted endoscopic surgery datasets EndoVis 2017~\cite{EndoVis_2017} (denoted as \evda), and  EndoVis 2018~\cite{EndoVis_2018} (denoted as \evdb) 
datasets for our experiments.
\begin{enumerate*}[label=\textbf{(\arabic*)}]
\item \evda dataset contains ten videos from the da-Vinci robotic system and provides annotations of 6 robotic instruments and an ultrasound probe. We adopted the 4-fold cross-validation from \cite{TernausNet} for fair comparison with 1800 frames ($8 \times 225$). The fold-wise split makes it 1350 and 450 frames for training and validation, respectively.
\item \evdb is a robotic instrument clinical dataset that includes organs and surgical items like gauze and suturing thread and contains the instrument super category but not the instrument type. This dataset is additionally annotated for instrument types by \cite{gonzalez2020isinet} with seven robotic instrument types and 11 training videos, and four testing videos with 149 images each. They provide the annotations as image pixels but do not provide instance labels. We annotated the instances ourselves for our experiments.
\end{enumerate*}

\mypara{Evaluation}
We categorize the compared methods into two categories. \evs methods are the ones that have reported their accuracy on \evda or \evdb datasets. \nli models are the instance segmentation methods proposed for natural images. For \evda and \evdb datasets, we evaluate the performance on the Challenge \iou (Ch\_\iou) metric as proposed in the \evda challenge~\cite{allan20192017} and ISINet \iou (ISI\_\iou) and mean class \iou (mc\iou) metrics proposed in~\cite{gonzalez2020isinet}. 

\section{Experiments and Results}

\subsection{Implementation Details}

\myfirstpara{Backbones}
%
The proposed \msma module can be added to any existing instance segmentation method as an additional stage. To validate this, we added \msma on two methods with very different architectures: a CNN-based MaskRCNN~\cite{MaskRCNN} and a newer transformer-based Mask2former~\cite{mask2former}. The latter is used for validation of third stage and performs poorly compared to the former. We report most of the results using the MaskRCNN as the initial stage. We refer to both the models (based on MaskRCNN or Mask2former) as \tss and explicitly specify the architecture type when using the transformer architecture.  

\mypara{Training}
We first train the first two stages using a regression loss, cross-entropy classification loss, and per-pixel segmentation loss.  
We have used an ImageNet pre-trained ResNet-50-FPN model to match the \sota ~\cite{gonzalez2020isinet} architecture and finetuned it for the instrument dataset. We resize each image to a size of $(1333,800)$. Stochastic Gradient Descent with a learning rate of $20^{-2}$ is used to train the two stages simultaneously for 12 epochs.

For stage 3 (\msma), we first use the pre-trained weights of ResNet-50-FPN from the box proposal module and freeze them initially to avoid over-fitting. The classification head of the \msma classifier is first trained using ground truth instance masks for ten epochs using cross-entropy-loss. Then it is trained for 15 epochs using arc loss. After 25 epochs, we unfreeze the weights of ResNet also and train the complete \msma classifier end-to-end for five epochs using arc loss. Finally, only the classification layer is trained using cross-entropy loss. For training the \msma classifier, we have resized each image to $(224,224)$. The ground truth masks were resized to $(56, 56)$ while masking the features to match the feature resolution of the last layer of block 3 in ResNet. 
The mask attended classification head of \msma is trained using Adam optimizer with a learning rate of $10^{-5}$, whereas the end-to-end training of \msma is done using a learning rate of $10^{-7}$.

\mypara{Inference}
During inference, we set the score threshold of 0.0 to accommodate all the classes and select only the top 5 instances because a typical frame contains approximately 3 to 4 instruments in the ground truth. 


\begin{table*}[t]
	\centering
	\resizebox{\textwidth}{!}{%
		\begin{tabular}{lllllllllllll}
			\hline
			\textbf{Method} &
			\textbf{Conference} &
			\textbf{Arch.} &
			\textbf{Ch\_} &
			\textbf{ISI\_} &
			\multicolumn{7}{c}{\textbf{Instrument Classes IOU}} &
			\textbf{mc} \\ 
			\cline{6-12}
			\textbf{} &
			\textbf{} &
			\textbf{} &
			\textbf{\iou} &
			\textbf{\iou} &
			\textbf{BF} &
			\textbf{PF} &
			\textbf{LND} &
			\textbf{VS/ SI} &
			\textbf{GR/ CA} &
			\textbf{MCS} &
			\textbf{UP} &
			\textbf{\iou} \\ 
			\hline
			\multicolumn{13}{c}{\textbf{Dataset \evda}} \\
			\hline
			\multicolumn{3}{l}{\textbf{\nli Methods}}    &       &       &       &       &       &       &       &       &       &       
			\\
			MaskRCNN~\cite{MaskRCNN} &
			ICCV17 & R50 & 45.65 & 41.77 & 27.59 & 33.67 & 43.96 & 17.95 & 0.80 & 4.20 & 8.98 & 19.59 
			\\
			CascadeRCNN~\cite{CascadeRCNN}                           & CVPR18    & R50    & 49.03 & 39.9  & 33.47 & 32.03 & 44.1  & 16.36 & 1.38  & 3.74  & 10.94 & 20.29 \\
			HTC~\cite{HTC}                                   & CVPR19    & R50    & 43.81 & 40.39 & 35.86 & 27.01 & 46.3  & 14.16 & 1.36  & 7.05  & 9.4   & 20.96 \\
			MScoring\_RCNN~\cite{MaskScoringRCNN}                      & CVPR19    & R50    & 47.63 & 44.54 & 37.95 & 38.48 & 49.43 & 13.55 & 2.57  & 3.93  & 9.52  & 25.23 \\
			SimCal~\cite{SimCal}                                & ECCV20    & R50    & 49.56 & 45.71 & 39.44 & 38.01 & 46.74 & 16.52 & 1.9   & 1.98  & 13.11 & 23.78 \\
			CondInst~\cite{CondInst}                               & ECCV20    & R50    & 59.02 & 52.12 & 44.29 & 38.03 & 47.38 & 24.77 & 4.51  & 15.21 & 15.67 & 27.12 \\
			BMaskRCNN~\cite{BMaskRCNN}                              & ECCV20    & R50    & 49.81 & 38.81 & 32.89 & 32.82 & 41.93 & 12.66 & 2.07  & 1.37  & 14.43 & 19.74 \\
			SOLO~\cite{SOLO}                                   & NeurIPS20 & R50    & 35.41 & 33.72 & 22.05 & 23.17 & 41.07 & 7.68  & 0     & 11.29 & 4.6   & 15.79 \\
			SCNet~\cite{SCNet}                                  & AAAI21    & R50    & 48.17 & 46.92 & 43.96 & 29.54 & 48.75 & 22.89 & 1.19  & 4.9   & 14.47 & 25.98 \\
			MFTA~\cite{MFTA}                                  & CVPR21    & R50    & 46.16 & 41.77 & 31.16 & 35.07 & 39.9  & 12.05 & 2.28  & 6.08  & 11.61 & 20.27 \\
			DetectoRS~\cite{DetectoRS}                             & CVPR21    & R50    & 50.93 & 47.38 & 48.54 & 34.36 & 49.72 & 20.33 & 2.04  & 8.92  & 10.58 & 24.93 \\
			Orienmask~\cite{Orienmask}                              & ICCV21    & Dknt53  & 42.09 & 39.27 & 40.42 & 28.78 & 44.48 & 12.11 & 3.91  & 15.18 & 12.32 & 23.22 \\
			QueryInst~\cite{QueryInst}                              & ICCV21    & R50    & 33.59 & 33.06 & 20.87 & 12.37 & 46.75 & 10.48 & 0.52  & 0.39  & 4.58  & 15.32 \\
			FASA~\cite{FASA}                                  & ICCV21    & R50    & 34.38 & 29.67 & 20.13 & 18.81 & 39.12 & 8.34  & 0.68  & 2.17  & 3.46  & 13.24 \\
			Mask2Former~\cite{mask2former}                                  & CVPR22    & Trfmr    & 40.39 & 39.84 & 19.60 & 20.22 & 45.44 & 11.95  & 0.00  & 1.48  & 22.10  & 17.78 \\
			\textbf{\tss (+Mask2former)}           &     & R50     & 53.31 & 51.2 & 49.48 & 29.91 & 70.61 & 32.98  & 19.53  & 18.35  & 49.51  & 38.13 \\ 
			\cline{1-13} 
			\multicolumn{3}{l}{\textbf{\evs Methods}}    &       &       &       &       &       &       &       &       &       &       \\
			TernausNet-11~\cite{TernausNet}                         & ICMLA18   & UNet11 & 35.27 & 12.67 & 13.45 & 12.39 & 20.51 & 5.97  & 1.08  & 1     & 16.76 & 10.17 \\
			MF-TAPNET~\cite{MFTAPNet}                             & MICCAI19  & UNet   & 37.35 & 13.49 & 16.39 & 14.11 & 19.01 & 8.11  & 0.31  & 4.09  & 13.4  & 10.77 \\
			ISINET~\cite{gonzalez2020isinet}                             & MICCAI20  & R50    & 55.62 & 52.2  & 38.7  & 38.5  & 50.09 & 27.43 & 2.01  & 28.72 & 12.56 & 28.96 \\
			TraSeTR~\cite{TraSeTR}                                & ICRA22    & Trfmr  & 60.4  & 65.2  & 45.2  & \textbf{56.7}  & 55.8  & \textbf{38.9}  & 11.4  & 31.3  & 18.2  & 36.79 \\
			\textbf{\tss (+MaskRCNN)}   &           & R50    & \textbf{72.54} & \textbf{71.99} & \textbf{75.08} & 54.32 & \textbf{61.84} & 35.5  & \textbf{27.47} & \textbf{43.23} & \textbf{28.38} & \textbf{46.55} \\ 
			\hline
			\multicolumn{13}{c}{\textbf{Dataset \evdb}} \\
			\hline
			\multicolumn{3}{l}{\textbf{\nli Methods}}    &       &       &       &       &       &       &       &       &       &       \\
			MaskRCNN~\cite{MaskRCNN}                           & ICCV17    & R50    & 69.41 & 67.94 & 72.85 & 43.13 & 0.85  & 32.63 & 0     & 86.16 & 0     & 33.66 \\
			CascadeRCNN~\cite{CascadeRCNN}                           & CVPR18    & R50    & 67.11 & 66.29 & 71.22 & 33.6  & 4.94  & 0     & 0     & 90.61 & 2.62  & 29    \\
			HTC~\cite{HTC}                                    & CVPR19    & R50    & 69.07 & 68.04 & 72.45 & 36.64 & 1.64  & 37.04 & 0     & 88.27 & 1.95  & 34    \\
			MScoring\_RCNN~\cite{MaskScoringRCNN}                      & CVPR19    & R50    & 65.19 & 64.04 & 68.69 & 31.23 & 4.81  & 0     & 0     & 88.23 & 1.75  & 27.82 \\
			SimCal~\cite{SimCal}                                & ECCV20    & R50    & 68.56 & 67.58 & 73.67 & 40.35 & 5.57  & 0     & 0     & 89.84 & 0     & 29.92 \\
			CondInst~\cite{CondInst}                               & ECCV20    & R50    & 72.27 & 71.55 & 77.42 & 37.43 & 7.77  & 43.62 & 0     & 87.8  & 0     & 36.29 \\
			BMaskRCNN~\cite{BMaskRCNN}                             & ECCV20    & R50    & 68.94 & 67.23 & 70.04 & 28.91 & 9.97  & 45.01 & 4.28  & 86.73 & 3.31  & 35.46 \\
			SOLO~\cite{SOLO}                                   & NeurIPS20 & R50    & 65.59 & 64.88 & 69.46 & 23.92 & 2.61  & 36.19 & 0     & 87.97 & 0     & 31.45 \\
			SCNet~\cite{SCNet}                                & AAAI21    & R50    & 71.74 & 70.99 & 78.4  & 47.97 & 5.22  & 29.52 & 0     & 86.69 & 0     & 35.4  \\
			MFTA~\cite{MFTA}                                  & CVPR21    & R50    & 69.2  & 67.97 & 71    & 31.62 & 3.93  & 43.48 & 9.9   & 87.77 & 3.86  & 35.94 \\
			DetectoRS~\cite{DetectoRS}                             & CVPR21    & R50    & 66.69 & 65.06 & 73.94 & 46.85 & 0     & 0     & 0     & 79.92 & 0     & 28.67 \\
			Orienmask~\cite{Orienmask}                              & ICCV21    & Dknt53  & 67.69 & 66.77 & 68.95 & 38.66 & 0     & 31.25 & 0     & 91.21 & 0     & 32.87 \\
			QueryInst~\cite{QueryInst}                             & ICCV21    & R50    & 66.44 & 65.82 & 74.13 & 31.68 & 2.3   & 0     & 0     & 87.28 & 0     & 27.91 \\
			FASA~\cite{FASA}                                   & ICCV21    & R50    & 68.31 & 66.84 & 72.82 & 37.64 & 5.62  & 0     & 0     & 89.02 & 1.03  & 29.45 \\
			Mask2Former~\cite{mask2former}                           & CVPR22    & Trfmr  & 65.47 & 64.69 & 69.35 & 24.13 & 0     & 0     & 0     & 89.96 & 10.29 & 27.67 \\
			\textbf{\tss (+Mask2former)} &           & R50    & 67.78 & 67.06 & 71.18 & 29.77 & 1.59  & 0     & 0     & 90.61 & 10.29 & 29.06 \\
			\cline{1-13} 
			\multicolumn{3}{l}{\textbf{\evs Methods}}    &       &       &       &       &       &       &       &       &       &       \\
			TernausNet-11~\cite{TernausNet}                          & ICMLA18   & UNet11 & 46.22 & 39.87 & 44.2  & 4.67  & 0     & 0     & 0     & 50.44 & 0     & 14.19 \\
			MF-TAPNET~\cite{MFTAPNet}                             & MICCAI19  & UNet   & 67.87 & 39.14 & 69.23 & 6.1   & 11.68 & 14    & 0.91  & 70.24 & 0.57  & 24.68 \\
			ISINET~\cite{gonzalez2020isinet}                                & MICCAI20  & R50    & 73.03 & 70.97 & 73.83 & 48.61 & 30.98 & 37.68 & 0     & 88.16 & 2.16  & 40.21 \\
			TraSeTR~\cite{TraSeTR}             & ICRA22    & Trfmr  & \textbf{76.2}  & \_  & 76.3  & ~\textbf{53.3}  & \textbf{46.5} & 40.6  &\textbf{13.9}  & 86.3  & \textbf{17.5}  & \textbf{47.77} \\
			\textbf{\tss (+MaskRCNN)}    &           & R50    & 75.81 & 74.02 & \textbf{77.22} & 50.87 & 19.83 & \textbf{50.59} & 0     & \textbf{92.12} & 7.44  & 42.58 \\ 
			\hline \\
		\end{tabular}%
	} \\
	\caption{Performance of \sota instance segmentation methods on \evda and \evdb instrument segmentation datasets. (R50 represents ResNet-50-FPN, Trfmr represents Transformer,  BF-Bipolar Forceps, PF-Prograsp Forceps, LND-Large Needle Driver, VS/SI- Vessel Sealer/ Suction Instrument, GR/CA- Grasping Retractor/Clip Applier, MCS-Monopolar Curved Scissors, UP-Ultrasound Probe)}
	\label{tab:comparison_EV17_18}
\end{table*}

\subsection{Analysis}

\myfirstpara{Comparison with \textbf{\sota}}
We compare \tss for instrument type segmentation with \evs methods that include semantic segmentation~\cite{TernausNet, MFTAPNet} and instance segmentation approaches~\cite{gonzalez2020isinet, TraSeTR}. For the \nli techniques, we use the source code provided by the authors to train the models for the mentioned datasets and use our inference parameters. We add region-based \nms as mentioned in ~\cref{sec:Our_Proposal} as a post-processing step on the predicted masks of an image and report the \iou scores after post-processing for all the models~(see \cref{tab:comparison_EV17_18}). 
\begin{figure*}
	\centering
	\includegraphics[width=0.9\linewidth]{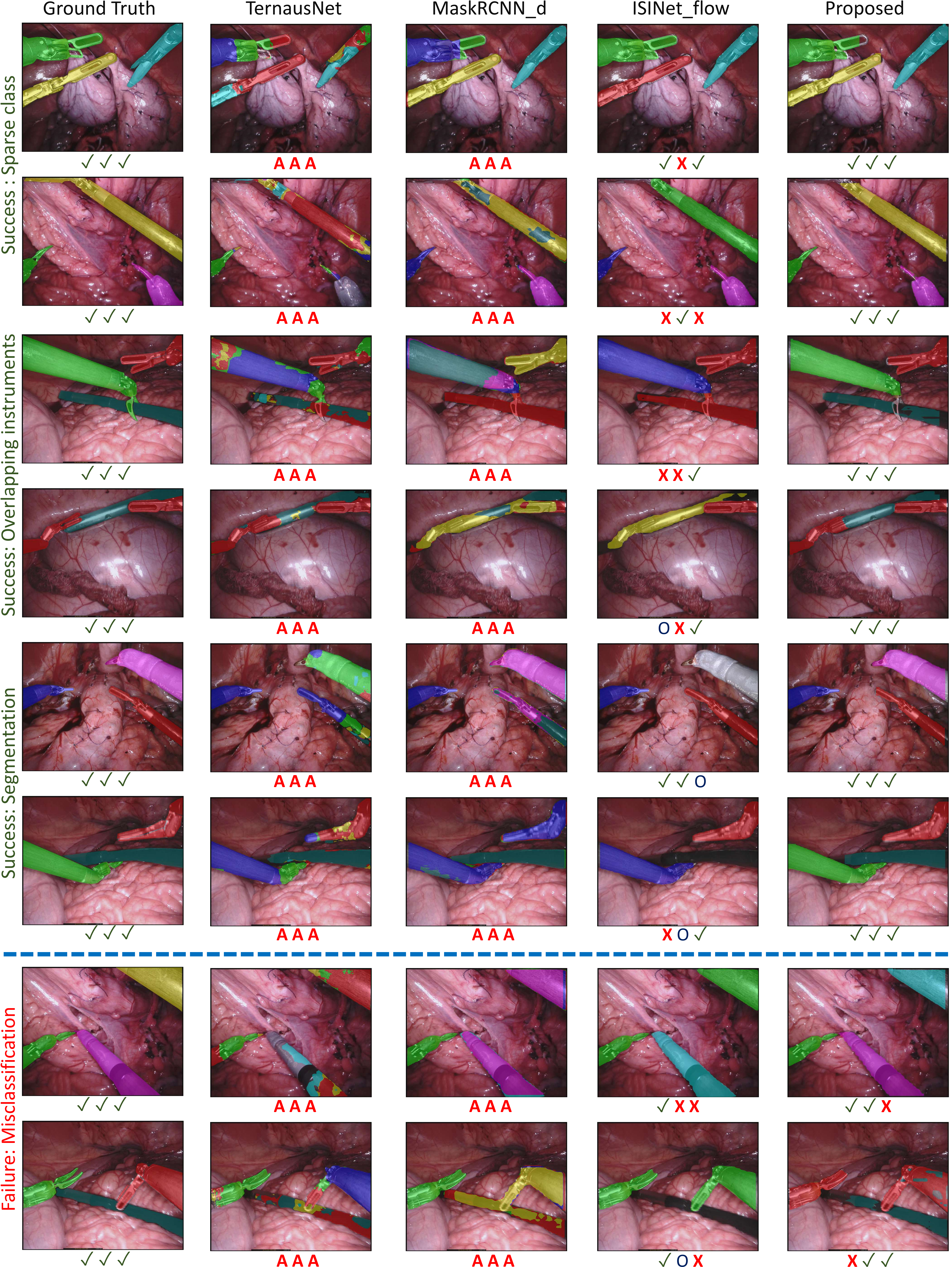}
	\caption{Qualitative Analysis for the comparison of instance segmentation: 4 symbols are used to show the results; \cmark~represents the instance labeled correctly, \xmark shows the misclassified instance, and `O' represents missed instance, A letter `A' indicates an ambiguous instance, where it is ambiguous to select the class of the instrument either due to over-segmentation or due to multiple copies of instrument classes at the same region. We show better classification in the cases of sparse class, overlapping instruments, and do not miss instrument instances. Our failure cases include cases where the instance shows the only shaft of the instrument and a significant change in instrument orientation.}
	\label{fig:QualitativeAnalysis}
\end{figure*}

For \evda, \tss outperforms all the \nli methods and the other \ evs-based instance segmentation methods. It improves over \isinet~\cite{gonzalez2020isinet} by 30\% Ch\_\iou and 60 \% mc\iou, showing that the mask-based classification using stage 3 improves the results by a considerable margin with only using the spatial information. Even though \trasetr~\cite{TraSeTR} explores transformer-based architecture with mask classification paradigm and also uses the temporal information, \tss outperforms it by a margin of 20\% on Ch\_\iou and 26\% on mc\iou achieving the \sota results. 
For \evdb dataset, \tss outperforms \isinet~\cite{gonzalez2020isinet} by a slight margin of 3.8\% on Ch\-\iou and 5.8\% on mc\iou. In comparison to \trasetr, the results are slightly lesser, which shows that for this dataset, apart from classification, other determining factors like temporal information of the instruments favored a tracking-based method.

\mypara{Validating Reasons for low accuracy of \sota}
%
In \cref{tab:comparison_EV17_18}, we compare \sota instance segmentation on \evs datasets. We have analyzed why the \nli models are less accurate. As noted before, we make three claims about \sota's low segmentation accuracy. First, the present two-stage classification heads are weak and are the bottleneck in accuracy. Second, the instrument's oblique posture and design allow the background to seep into the rectangular boxes, complicating classification. Third, cross entropy-based loss makes learning visually similar instruments harder. We investigated all three claims.

For the first claim, we replace a two-stage model's (MaskRCNN) predicted label with the ground truth label. This simple change increases the mask AP50 score from 0.65 to 0.90, showing classification inaccuracy.

We evaluate Video 1 frames of \evda containing ultrasound probes for the second claim. MaskRCNN predicts 37 ultrasound probe bounding boxes out of 224, with 26 having an \iou of 0.75. 22 of the 26 boxes had an aspect ratio greater than 3, whereas just 4 had an aspect ratio less than 3. Elongated boxes' prediction accuracy was 84\%. MaskRCNN has substantially greater accuracy when the ground truth box firmly hugs the instrument.

For the third claim, we compare the outcomes after training the third stage using cross-entropy loss and arc loss-based metric learning.

\subsection{Qualitative Analysis}

The qualitative classification accuracy results are shown in \cref{fig:QualitativeAnalysis}. We show the comparative results of a semantic segmentation method, a typical instance segmentation method, and an \evs method qualitatively. We show better classification in the cases of sparse class and overlapping instruments. Our failure cases include cases where the instance shows only the instrument shaft and a significant change in instrument orientation. Our better performance on the sparse class of Grasping retractor is due to the metric learning-based training loss. Because we devise a mask-attention-based classifier, the network performs well in classifying overlapping regions. We only focused on improving the classification of instance segmentation and not on the temporal context because of the high dependency of the classification accuracy for the next stage of applications. A future direction of this problem can be towards improving the masks and improving the instance labels further based on the temporal information.


\begin{table}[t]
\centering
\resizebox{0.9\linewidth}{!}{%
\begin{tabular}{llll}
\hline
     & \textbf{Model}  & \textbf{Ch\_ \iou} \\ \hline
Stage 1 \& 2 & Stage 1 \& 2     &    37.97    \\
     & Stage 1 \& 2\_wsr &      53.30           \\
     & Stage 1 \& 2\_maskc &    57.35        \\
     & Stage 1 \& 2\_wma &     57.09 \\
Stage 3 & Stage 3\_cel &    63.63   \\
    & \tss &      72.54 \\
     \hline
\end{tabular}%
}
\caption{Ablation Studies of the proposed \tss on \evda}
\label{tab:Ablation}
\end{table}



\subsection{Ablation studies}
\label{sec:Ablation studies}

\cref{tab:Ablation} gives the result of various ablation studies performed to understand the importance of various modules in our system. We describe the notation below:

\mypara{Stage 1 \& 2}
Here, we report the accuracy obtained by our model after 2nd stage without the post-processing. Since our model uses the MaskRCNN backbone for the first two stages, this is essentially the accuracy of MaskRCNN using our hyper-parameters. 

\mypara{Stage 1 \& 2 \_wsr} 
Result of the second stage using our post-processing of non-maximal suppression across classes.
  
\mypara{Stage 1 \& 2 \_maskc} 
In the current MaskRCNN, classification and mask prediction are performed in parallel. As per the thesis of this paper, the classification of erroneous boxes is fragile. Hence, in this experiment, we have changed the ordering of the stage 2 predictions. Now the classification is not performed parallel to mask prediction, but after the mask prediction and is done on the mask attended features.

\mypara{Stage 1 \& 2 \_wma} 
We have explored whether we can use features of stage 2 for classification instead of training a separate classifier stage. In this experiment, we keep the first two stages as is, but after stage 2, use the mask attended features from stage 2 only. The difference between this and the previous configuration is that in the previous config, the original classifier of MaskRCNN was disabled, but in this one, it remains as is.

\mypara{Stage 3 \_cel} 
Here we train S3Net third stage-trained using cross-entropy instead of arc loss. The lower accuracy of this configuration serves to validate one of this paper's key observations, that the instrument's visual similarity makes it difficult to classify, and hence learning representation and classification should be disentangled using metric learning.

\section{Conclusion}
\label{sec:Conclusion}

In this study, we investigated the reasons for the low performance of techniques developed for natural image, on the surgical instrument segmentation tasks. We also showed how carefully designed architectural innovations can successfully mitigate the challenges. We conduct exhaustive experiments on the benchmark robot-assisted surgery datasets EndoVis2017 (\evda), and EndoVis2018 (\evdb). The proposed method generalizes well on all these datasets, outperforms the instance segmentation methods with varying backbones, and achieves at least 12 points (20\%) improvement over the \sota on the benchmark \evda challenge. We conclude that adding a third classification stage improves the results for applications involving fine-grained classification, such as surgical tool segmentation. We hope that our analysis and the innovations to mitigate the challenges specific to surgical instruments will spark similar interest among researchers for the effective application of advancements in natural imaging models to surgical imaging tasks. The proposed framework can be used for downstream applications that depend on tool identification and segmentation. We plan to extend the method to include tracking cues and further improve classification accuracy. 
%



\mypara{Acknowledgements}
This work was supported by Department of Biotechnology, Ministry of Science and Technology, India (Project No. BT/PR13455/CoE/34/24/2015)

{\small
\bibliographystyle{ieee_fullname}
\bibliography{egbib}
}

\appendix
\begin{table*}[t]
	\centering
	\resizebox{\textwidth}{!}{%
		\begin{tabular}{lllllllllllll}
			\hline
			\textbf{Method} &
			\textbf{Conference} &
			\textbf{Arch.} &
			\textbf{Ch\_} &
			\textbf{ISI\_} &
			\multicolumn{7}{c}{\textbf{Instrument Classes IOU}} &
			\textbf{mc} \\ 
			\cline{6-12}
			\textbf{} &
			\textbf{} &
			\textbf{} &
			\textbf{\iou} &
			\textbf{\iou} &
			\textbf{BF} &
			\textbf{PF} &
			\textbf{LND} &
			\textbf{SI} &
			\textbf{CA} &
			\textbf{MCS} &
			\textbf{UP} &
			\textbf{\iou} \\ 
			\hline
			\multicolumn{13}{c}{\textbf{Dataset \evdb with val\_old}} \\
			\hline
			\multicolumn{3}{l}{\textbf{\nli Methods}}    &       &       &       &       &       &       &       &       &       &       
			\\
			MaskRCNN~\cite{MaskRCNN} &
			ICCV17 & R50 & 69.06 &	66.79&	72.85	&36.80	&0.55&	32.63&	0&	86.16	&0&	32.71
			\\
			CascadeRCNN~\cite{CascadeRCNN}                           & CVPR18    & R50    & 67.77&	66.37&	71.22	&35.84	&7.93&	0	&0	&90.61&	2.62	&29.75 \\
			HTC~\cite{HTC}                                   & CVPR19    & R50    & 69.49	&67.70&	72.45&	36.71	&4.05	&37.04&	0	&88.27	&1.95	&34.35 \\
			MScoring\_RCNN~\cite{MaskScoringRCNN}                      & CVPR19    & R50    & 65.46	&63.73&	68.69	&31.20	&3.52	&0&	0	&88.23&	1.75	&27.63 \\
			SimCal~\cite{SimCal}                                & ECCV20    & R50    & 69.05	&67.25	&73.67&	36.96	&9.34	&0	&0	&89.84	&0	&29.97 \\
			CondInst~\cite{CondInst}                               & ECCV20    & R50    & 72.83	&71.12	&77.42&	40.24&	5.05	&43.62	&0	&87.80	&0	&36.31 \\
			BMaskRCNN~\cite{BMaskRCNN}                              & ECCV20    & R50    & 70.24&	67.82	&70.04	&34.59&	15.19&	45.01&	4.28&	86.73&	3.31&	37.02 \\
			SOLO~\cite{SOLO}                                   & NeurIPS20 & R50    & 66.79&	65.64	&69.46	&33.63	&3.75	&36.19	&0	&87.97	&0	&33.00 \\
			SCNet~\cite{SCNet}                                  & AAAI21    & R50    & 71.31	&69.43&	78.40	&34.09&	3.76	&29.52	&0	&86.69	&0	&33.21\\
			MFTA~\cite{MFTA}                                  & CVPR21    & R50    & 69.63	&67.73&	71.00	&34.99	&2.25	&43.48	&9.90	&87.77	&3.86	&36.18 \\
			DetectoRS~\cite{DetectoRS}                             & CVPR21    & R50    & 65.77	&63.31	&73.94	&32.36	&0	&0	&0	&79.92	&0	&26.60 \\
			Orienmask~\cite{Orienmask}                              & ICCV21    & Dknt53  & 67.47&	65.66&	68.95	&29.90	&0.30	&31.25	&0	&91.21	&0	&31.66 \\
			QueryInst~\cite{QueryInst}                              & ICCV21    & R50    & 66.72	&65.75	&74.13	&34.88&	2.51&	0&	0&	87.28&	0&	28.40 \\
			FASA~\cite{FASA}                                  & ICCV21    & R50    & 68.69&	66.28&	72.82	&30.53	&11.07&	1.03&	0&	89.02&	0 &	29.21\\
			Mask2Former~\cite{mask2former}                                  & CVPR22    & Trfmr    & 65.15	&64.25	&69.35&	23.46&	0	&0	&0	&89.96	&10.29	&27.58 \\
			\textbf{\tss (+Mask2former)}           &     & R50     & 67.45&	66.62	&71.18	&31.26	&1.56	&0	&0	&90.61	&10.29	&29.27 \\ 
			\cline{1-13} 
			\multicolumn{3}{l}{\textbf{\evs Methods}}    &       &       &       &       &       &       &       &       &       &       \\
			TernausNet-11~\cite{TernausNet}                          & ICMLA18   & UNet11 & 46.22 & 39.87 & 44.2  & 4.67  & 0     & 0     & 0     & 50.44 & 0     & 14.19 \\
			MF-TAPNET~\cite{MFTAPNet}                             & MICCAI19  & UNet   & 67.87 & 39.14 & 69.23 & 6.1   & 11.68 & 14    & 0.91  & 70.24 & 0.57  & 24.68 \\
			ISINET~\cite{gonzalez2020isinet}                                & MICCAI20  & R50    & 73.03 & 70.97 & 73.83 & 48.61 & 30.98 & 37.68 & 0     & 88.16 & 2.16  & 40.21 \\
			TraSeTR~\cite{TraSeTR}             & ICRA22    & Trfmr  & \textbf{76.2}  & \_  & 76.3  & ~\textbf{53.3}  & \textbf{46.5} & 40.6  &\textbf{13.9}  & 86.3  & \textbf{17.5}  & \textbf{47.77} \\
			\textbf{\tss (+MaskRCNN)}    &           & R50    & 75.81 & 74.02 & \textbf{77.22} & 50.87 & 19.83 & \textbf{50.59} & 0     & \textbf{92.12} & 7.44  & 42.58 \\ 
			\hline
			\multicolumn{13}{c}{\textbf{Dataset \evdb with val\_new}} \\
			\hline
   
			\multicolumn{3}{l}{\textbf{\nli Methods}}    &       &       &       &       &       &       &       &       &       &       \\
			MaskRCNN~\cite{MaskRCNN}                           & ICCV17    & R50    & 69.41 & 67.94 & 72.85 & 43.13 & 0.85  & 32.63 & 0     & 86.16 & 0     & 33.66 \\
			CascadeRCNN~\cite{CascadeRCNN}                           & CVPR18    & R50    & 67.11 & 66.29 & 71.22 & 33.6  & 4.94  & 0     & 0     & 90.61 & 2.62  & 29    \\
			HTC~\cite{HTC}                                    & CVPR19    & R50    & 69.07 & 68.04 & 72.45 & 36.64 & 1.64  & 37.04 & 0     & 88.27 & 1.95  & 34    \\
			MScoring\_RCNN~\cite{MaskScoringRCNN}                      & CVPR19    & R50    & 65.19 & 64.04 & 68.69 & 31.23 & 4.81  & 0     & 0     & 88.23 & 1.75  & 27.82 \\
			SimCal~\cite{SimCal}                                & ECCV20    & R50    & 68.56 & 67.58 & 73.67 & 40.35 & 5.57  & 0     & 0     & 89.84 & 0     & 29.92 \\
			CondInst~\cite{CondInst}                               & ECCV20    & R50    & 72.27 & 71.55 & 77.42 & 37.43 & 7.77  & 43.62 & 0     & 87.8  & 0     & 36.29 \\
			BMaskRCNN~\cite{BMaskRCNN}                             & ECCV20    & R50    & 68.94 & 67.23 & 70.04 & 28.91 & 9.97  & 45.01 & 4.28  & 86.73 & 3.31  & 35.46 \\
			SOLO~\cite{SOLO}                                   & NeurIPS20 & R50    & 65.59 & 64.88 & 69.46 & 23.92 & 2.61  & 36.19 & 0     & 87.97 & 0     & 31.45 \\
			SCNet~\cite{SCNet}                                & AAAI21    & R50    & 71.74 & 70.99 & 78.4  & 47.97 & 5.22  & 29.52 & 0     & 86.69 & 0     & 35.4  \\
			MFTA~\cite{MFTA}                                  & CVPR21    & R50    & 69.2  & 67.97 & 71    & 31.62 & 3.93  & 43.48 & 9.9   & 87.77 & 3.86  & 35.94 \\
			DetectoRS~\cite{DetectoRS}                             & CVPR21    & R50    & 66.69 & 65.06 & 73.94 & 46.85 & 0     & 0     & 0     & 79.92 & 0     & 28.67 \\
			Orienmask~\cite{Orienmask}                              & ICCV21    & Dknt53  & 67.69 & 66.77 & 68.95 & 38.66 & 0     & 31.25 & 0     & 91.21 & 0     & 32.87 \\
			QueryInst~\cite{QueryInst}                             & ICCV21    & R50    & 66.44 & 65.82 & 74.13 & 31.68 & 2.3   & 0     & 0     & 87.28 & 0     & 27.91 \\
			FASA~\cite{FASA}                                   & ICCV21    & R50    & 68.31 & 66.84 & 72.82 & 37.64 & 5.62  & 0     & 0     & 89.02 & 1.03  & 29.45 \\
			Mask2Former~\cite{mask2former}                           & CVPR22    & Trfmr  & 65.47 & 64.69 & 69.35 & 24.13 & 0     & 0     & 0     & 89.96 & 10.29 & 27.67 \\
			\textbf{\tss (+Mask2former)} &           & R50    & 67.78 & 67.06 & 71.18 & 29.77 & 1.59  & 0     & 0     & 90.61 & 10.29 & 29.06 \\
			\cline{1-13} 
			\multicolumn{3}{l}{\textbf{\evs Methods}}    &       &       &       &       &       &       &       &       &       &       \\
			\textbf{\tss (+MaskRCNN)}    &           & R50    & 74.00	&73.73	&77.22	&39.54	&16.96	&50.59	&0&	92.12	&7.45&	40.56 \\ 
			\hline \\
		\end{tabular}%
	} \\
	\caption{Performance of \sota instance segmentation methods on \evdb val\_old and \evdb val\_new instrument segmentation datasets. (R50 represents ResNet-50-FPN, Trfmr represents Transformer,  BF-Bipolar Forceps, PF-Prograsp Forceps, LND-Large Needle Driver, SI-  Suction Instrument, CA- Clip Applier, MCS-Monopolar Curved Scissors, UP-Ultrasound Probe)}
	\label{tab:comparison_EV17_18}
\end{table*}

\section{Supplementary Material}

The sequence 2 in Endovis 2018 has a misclassification that we identified for the test dataset of the ground truth. The class 2 Prograsp Forceps is misclassified as class 3 Large Needle Driver by~\cite{gonzalez2020isinet} et al. This may have happened due to the tip being occluded by the tissue. We have updated the validation set in our released \evdb dataset with a val\_new dataset.

Also, to become consistent with the comparison with validation, we are providing an extension to the~\cref{tab:comparison_EV17_18} as below.

\end{document}